\documentclass{article}

\usepackage{PRIMEarxiv}

\usepackage[utf8]{inputenc}
\usepackage[T1]{fontenc}
\usepackage{hyperref}
\usepackage{url}
\usepackage{booktabs}
\usepackage{amsfonts}
\usepackage{nicefrac}
\usepackage{microtype}
\usepackage{graphicx}
\usepackage{multirow}
\usepackage{array}
\usepackage{siunitx}
\usepackage{xcolor}
\usepackage{listings}

\definecolor{codegreen}{rgb}{0,0.6,0}
\definecolor{codegray}{rgb}{0.5,0.5,0.5}
\definecolor{codepurple}{rgb}{0.58,0,0.82}
\definecolor{backcolour}{rgb}{0.97,0.97,0.97}

\title{\includegraphics[width=0.99\textwidth]{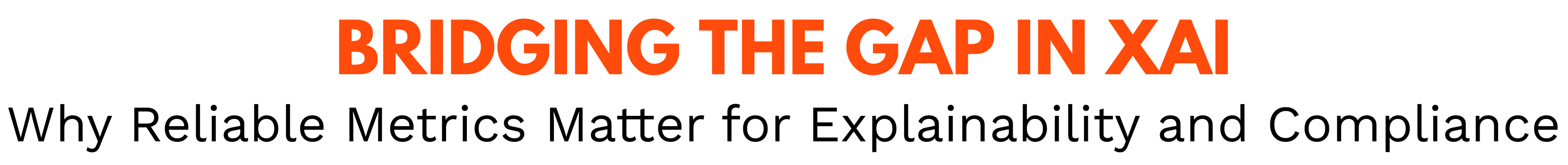}}

\author{
  Pratinav Seth, Vinay Kumar Sankarapu \\
  \affiliation{Lexsi Labs}\\
  \{pratinav.seth,v.k\}@lexsi.ai \\
}
\setabstract{%
Reliable explainability is not only a technical goal but also a cornerstone of private AI governance. As AI models enter high-stakes sectors, private actors such as auditors, insurers, certification bodies, and procurement agencies require standardized evaluation metrics to assess trustworthiness.
However, current XAI evaluation metrics remain fragmented and prone to manipulation, which undermines accountability and compliance. We argue that standardized metrics can function as \emph{governance primitives}, embedding auditability and accountability within AI systems for effective private oversight. Building upon prior work in XAI benchmarking, we identify key limitations in ensuring faithfulness, tamper resistance, and regulatory alignment.
Furthermore, interpretability can directly support model alignment by providing a verifiable means of ensuring behavioral integrity in General Purpose AI (GPAI) systems. This connection between interpretability and alignment positions XAI metrics as both technical and regulatory instruments that help prevent \emph{alignment faking}, a growing concern among oversight bodies.
We propose a \textsc{Governance-by -Metrics} paradigm that treats explainability evaluation as a central mechanism of private AI governance. Our framework introduces a hierarchical model linking transparency, tamper resistance, scalability, and legal alignment, extending evaluation from model introspection toward systemic accountability. Through conceptual synthesis and alignment with governance standards, we outline a roadmap for integrating explainability metrics into continuous AI assurance pipelines that serve both private oversight and regulatory needs.

}

\setkeywords{Explainable AI, Evaluation Metrics, General Purpose AI (GPAI), Interpretability, Model Alignment, Alignment Faking, Auditability, Accountability, Tamper Resistance, Regulatory Compliance, AI Assurance, Governance-by-Metrics}
\runningtitle{Bridging the Gap in XAI-Why Reliable Metrics Matter for Explainability and Compliance}

\begin{document}
\maketitle

\section{Introduction}
As AI systems evolve into generative and agentic architectures, the reliability of explainability metrics becomes a governance issue determining whether systems can be audited, trusted, and lawfully deployed. AI is already embedded in daily life and high-stakes domains \cite{chatgpt,Touvron2023LLaMAOA}, with applications spanning healthcare, finance, and law enforcement. As its impact grows, systems must be transparent and their decision-making explainable.
We argue XAI metrics are no longer only diagnostic but also enforcement levers governing transparency, trust, and accountability across the lifecycle, helping identify and mitigate risks \cite{Amodei2016ConcretePI, doi:10.1073/pnas.1611835114, Leike2018ScalableAA}. Yet a critical gap persists: \textbf{we lack standardized, reliable metrics to evaluate the effectiveness and trustworthiness of explanations}.

The evaluation landscape is fragmented and often subjective \cite{madsen2024interpretabilityneedsnewparadigm}, enabling manipulation and weakening comparisons across tasks \cite{wickstrom2024flexibilitymanipulationslipperyslope,hedstrom2023quantusexplainableaitoolkit}. To be effective in high-risk settings, metrics must reliably assess fidelity, robustness, and usability. While regulatory frameworks like the EU AI Act \cite{europaRegulation20241689,10489989} and ISO 42001 \cite{iso42001} provide legal baselines, effective AI governance also depends on private oversight mechanisms technical and institutional processes within organizations that continuously monitor and verify compliance. This aligns with emerging perspectives on private AI governance, where assurance bodies, insurers, and certification consortia use quantitative evaluation for continuous oversight. XAI metrics can anchor these mechanisms by providing verifiable signals of explainability quality for use in insurance underwriting, procurement, and certification.

This challenge extends to General Purpose AI (GPAI) systems, where interpretability-based alignment has shown measurable potential \cite{malik2025interpretabilityawarepruningefficientmedical,interpretability_guided_alignment_2025}. These findings suggest that explainability metrics can move beyond evaluation toward behavioral steering a crucial capability for both private assurance and regulatory compliance. As GPAI systems become more complex and autonomous, the need for governance-aligned interpretability becomes critical for ensuring these systems remain accountable and controllable.

\paragraph{Position:} Reliable XAI metrics are essential for technical progress and both private and regulatory compliance. Advanced systems particularly General Purpose AI (GPAI) models demand scalable, manipulation-resistant evaluation aligned with real-world governance needs. We call for metrics that are contextually adaptive and serve as instruments for private oversight. By building robust core metrics and aligning them with both regulatory frameworks and private governance mechanisms, we can establish consistent, useful evaluation. Collaboration among researchers, industry, and regulators is key to achieving meaningful, trustworthy, and compliant explanations.

The remainder of this paper is structured as follows: Section 2 provides background on XAI methods and evaluation metrics; Section 3 identifies key challenges; Section 4 examines mechanistic interpretability advances; Sections 5-6 establish requirements and alternative views; Section 7 outlines our governance roadmap; Section 8 discusses policy integration; Section 9 concludes; and Section 10 provides an impact statement.

\section{Background and Context}
Explainable AI (XAI) aims to make complex ML models understandable \cite{arrieta2019explainableartificialintelligencexai}, which is vital in high-impact domains like healthcare, finance, and law enforcement. Beyond accuracy, systems must provide transparent, accountable, and comprehensible explanations especially where outcomes are consequential \cite{jia2022roleexplainabilityassuringsafety}.

\textsc{Interpretability, explainability, and feature attribution} serve related but distinct goals \cite{Doshi-Velez2017a}. Interpretability concerns how readily humans can follow a model's reasoning; explainability provides reasons for predictions (e.g., LIME \cite{Ribeiro2016WhySI}, SHAP \cite{Lundberg2017AUA}); attribution quantifies each input's contribution to an output. Together, they advance accountability and trust.

\subsection{Explainability Methods}

\begin{itemize}
    \item \textsc{Intrinsic explainability} \cite{Lipton2018,Rudin2019} uses inherently interpretable models (e.g., decision trees, linear regression, rule-based systems). These expose decision logic directly but may trade accuracy for simplicity. Recent work questions the faithfulness of some "inherently interpretable" claims \cite{Jacovi2020}, and attention-based explanations have faced criticism \cite{Serrano2019,madsen2024interpretabilityneedsnewparadigm}.
    \item \textsc{Post hoc explainability} \cite{Lipton2018,Madsen2021} explains trained models without changing them. Notable methods include SHAP \cite{Lundberg2017AUA}, LIME \cite{Ribeiro2016WhySI}, Grad-CAM \cite{Selvaraju2016GradCAMVE}, and Integrated Gradients \cite{Sundararajan2017AxiomaticAF}. These methods are flexible and widely used but approximate model behavior.
    \item \textsc{Example-based methods} compare similar cases; mechanistic interpretability reverse-engineers internals \cite{olah2020zoom,Nanda2023ProgressMF}.
\end{itemize}

These approaches face limitations: approximations can miss true behavior; explanations may be inconsistent under small input changes; and they can be manipulated adversarially \cite{slack2020foolinglimeshapadversarial}. Moreover, with no control over the model, faithfulness can be difficult \cite{madsen2024interpretabilityneedsnewparadigm}.

\subsection{The Role of Evaluation Metrics in XAI}
While explainability methods have advanced considerably, their evaluation remains inconsistent, hindering reproducibility and comparability. Recent progress in quantitative analysis introduced a broad set of evaluation metrics \cite{agarwal2022openxai} for assessing reliability and effectiveness \cite{inproceedingxaimetrs}. These help researchers and practitioners assess how well explanations reflect model decision-making and meet requirements like transparency, robustness, and usability. Private governance initiatives increasingly rely on measurable indicators to inform assurance practices analogous to how financial auditing depends on standardized accounting principles. Reliable XAI metrics could thus serve as "auditing primitives" for model interpretability and robustness.
Over time, it has become clear that most XAI metrics can be grouped into six categories:

\begin{enumerate}
    \item \textbf{Faithfulness:} Metrics measure how well explanations reflect the model's true decision-making process, ensuring accuracy and alignment with actual predictions \cite{Bhatt2020EvaluatingAA,AlvarezMelis2018TowardsRI}.
    \item \textbf{Robustness:} Metrics evaluate stability under varying inputs, including adversarial attacks and perturbations, ensuring explanations maintain integrity across test conditions \cite{yeh2019infidelitysensitivityexplanations,agarwal2022rethinkingstabilityattributionbasedexplanations}.
    \item \textbf{Localisation:} Metrics assess the ability to highlight relevant regions or features that most influence model decisions, particularly important for image data \cite{kohlbrenner2020bestpracticeexplainingneural,Arras_2022}.
    \item \textbf{Complexity:} Metrics evaluate simplicity and comprehensibility, ensuring explanations are accessible to end-users without unnecessary complexity \cite{chalasani2020conciseexplanationsneuralnetworks}.
    \item \textbf{Randomisation (Sensitivity):} Metrics examine sensitivity to input data or parameter changes, ensuring explanations don't rely on trivial variations \cite{adebayo2020sanitycheckssaliencymaps,Hedstrm2024SanityCR}.
    \item \textbf{Axiomatic:} Metrics evaluate inherent properties like consistency, completeness, and preservation across architectures, grounded in theoretical foundations \cite{Sundararajan2017AxiomaticAF,kindermans2017unreliabilitysaliencymethods}.
\end{enumerate}
\subsection{Existing Evaluation Frameworks and Benchmarking Tools for Model Explainability}

Current evaluation frameworks face significant issues. Many metrics don't capture real-world model complexity, and benchmarks lack flexibility across domains. Research suggests shifting focus toward robustness, generalizability, and actionability, while accounting for model evolution.

M$^4$ Benchmark \cite{li2023mathcalm} and OpenXAI \cite{agarwal2024openxaitransparentevaluationmodel} address some gaps but have limitations: M$^4$ focuses on faithfulness without robustness, OpenXAI relies on synthetic data, and Quantus \cite{hedstrom2023quantusexplainableaitoolkit} struggles with human judgment alignment. Other tools like Captum \cite{kokhlikyan2020captumunifiedgenericmodel} focus on fairness and attribution but lack standardized comparison methods.

Specialized libraries like Ferret \cite{attanasio-etal-2023-ferret} and Inseq \cite{Sarti_2023} offer contributions: Ferret examines post-hoc methods but is limited to text models, while Inseq targets NLP sequence generation tasks.

\subsection{Distinctive Contributions and Novelty}
Unlike prior calls such as M$^4$~\cite{li2023mathcalm} and OpenXAI~\cite{agarwal2024openxaitransparentevaluationmodel}, which emphasize benchmark standardization, our position contributes three new governance-oriented dimensions:

\begin{enumerate}
    \item \textbf{Tamper-Resistance:} We propose manipulation-resilient metrics that decouple evaluation hyperparameters from model-specific artifacts, establishing auditability.
    \item \textbf{Regulatory Alignment:} We explicitly map metric requirements to global governance frameworks (EU AI Act, NIST AI RMF~\cite{nist2023ai}, ISO/IEC 42001~\cite{iso42001}).
    \item \textbf{Cross-Modality Integration:} We extend reliability evaluation to multimodal and agentic systems, capturing decision-making dependencies across modalities and agents.
    \item \textbf{Private Governance Integration:} We position XAI metrics as quantitative instruments for private oversight enabling certification, liability assessment, and procurement-based governance of AI systems, complementing statutory regulation.
\end{enumerate}

By embedding governance principles into metric design, this framework transforms evaluation from passive assessment to active oversight, offering a distinct governance-by-design pathway that serves both private and public governance needs.
With this foundation established, we next examine the critical challenges that hinder current XAI evaluation practices.

\section{Challenges in XAI Metrics}
Evaluating XAI methods presents several challenges that hinder reliability and adoption. These issues prevent effective, universally applicable evaluation frameworks needed by regulators, risk managers, and users. Major problems include fragmentation, subjectivity, and manipulation vulnerabilities. Heavy reliance on hyperparameters creates opportunities to tune for favorable results \cite{wickstrom2024flexibilitymanipulationslipperyslope,hoffman2019metricsexplainableaichallenges}.

\subsection{Neglect of Modern AI Models}
Current XAI evaluation metrics struggle to capture the complexity of modern AI models, particularly large language models and autoregressive systems. These models rely on intricate decision-making processes and large-scale architectures, requiring more adaptable evaluation methods.

Explainability research has mainly focused on image and tabular modalities, with recent efforts extending to NLP. However, multi-modal AI systems are becoming increasingly common, yet most XAI methods remain single-modality focused, limiting their applicability to models processing text, images, and structured data together. Among post-hoc explanation methods, only a few, such as Layer-wise Relevance Propagation \cite{achtibat2024attnlrpattentionawarelayerwiserelevance} and DLBacktrace \cite{sankarapu2024dlbacktracemodelagnosticexplainability}, extend to multi-modal settings, leaving a significant evaluation gap.

As multi-modal AI adoption grows, the lack of standardized evaluation frameworks hinders interpretability and trustworthiness across domains. Existing XAI methods fail to capture dependencies between modalities. For instance, vision-based methods like Grad-CAM fail to explain text contributions in vision-language models like CLIP, while SHAP and LIME overlook image-based reasoning. This limitation is particularly critical in applications such as medical AI, where decisions rely on both textual reports and diagnostic images. Without dedicated multi-modal explainability metrics, evaluating these models remains inconsistent and unreliable. Addressing this challenge requires new faithfulness metrics that measure explanation alignment across modalities, along with benchmarking datasets to establish industry-wide standards.

\subsection{Fragmentation and Manipulation Risks}
A key challenge is the fragmentation in XAI evaluation due to the lack of standardized frameworks. Different metrics are used across studies, making comparing results and drawing broad conclusions hard. This inconsistency slows progress and hinders the development of best practices. Without a unified system, the field lacks direction, limiting collaboration and advancements in XAI. This fragmentation mirrors the broader governance challenge without standardized metrics, explainability cannot serve as an accountability instrument.

XAI metrics are vulnerable to intentional or unintentional manipulation, which undermines trust in XAI systems and diminishes their practical value. Examples include adjusting evaluation parameters to achieve desired outcomes, optimizing explanations to perform well on specific metrics while not accurately reflecting the model's true decision-making process, and using adversarial inputs to create explanations that seem robust but fail in real-world applications. Without tamper-resistant benchmarks, evaluation risks devolving into \emph{explanation theater}, weakening both private and public trust.
As recent studies show, the sensitivity of XAI evaluations to hyperparameters whether tied to model architecture, attribution baselines, or metric parameters creates pathways for manipulation. For example, baseline selection in Integrated Gradients or perturbation order in faithfulness metrics can drastically shift rankings \cite{sturmfels2020visualizing,Sundararajan2017AxiomaticAF}. This interdependence highlights the need for tamper-resistant evaluation frameworks capable of maintaining metric stability under parameter variation.

Recent research has increasingly focused on the role of hyperparameters in XAI evaluations and how they can introduce confounding effects \cite{hedstrom2023quantusexplainableaitoolkit}. Studies have explored how sensitive attribution methods are to explanation hyperparameters like random seed or sample size \cite{bansal2020}, how baseline choices impact explanation outcomes \cite{sturmfels2020visualizing, Sundararajan2017AxiomaticAF}, and how model changes (optimizer, activation, learning rate, data splits) influence explanations \cite{karimi2023on}. Research has also investigated how normalization, randomization order, and similarity measures affect evaluation outcomes \cite{sanity2024, rong2022consistentefficientevaluationstrategy}.

To address these challenges and establish reliable governance-aligned metrics, we must first examine the technical foundation provided by recent advances in mechanistic interpretability. These developments demonstrate how interpretability can move beyond passive explanation toward active alignment and governance.

\section{Mechanistic Interpretability and Alignment for Governance}

Recent interpretability efforts such as TransformerLens~\cite{nanda2022transformerlens}, CircuitsVis~\cite{cooney2022circuitsvis}, and Scaling Monosemanticity~\cite{templeton2024scaling} reveal neuron-level insights into large models, yet they lack measurable governance alignment. Recent advances~\cite{malik2025interpretabilityawarepruningefficientmedical,interpretability_guided_alignment_2025} demonstrate that interpretability can \emph{directly guide model alignment}, showing that pruning and attribution-aware feedback can enhance both efficiency and ethical behavior. Such interpretability-guided alignment not only enhances model integrity but also provides measurable assurance to private auditors verifying alignment claims. This evidence supports interpretability as a concrete mechanism for aligning General Purpose AI (GPAI) systems under regulatory scrutiny.

Regulators have increasingly cited "alignment faking" models simulating ethical behavior without genuine internal compliance as a critical governance risk \cite{Lindstrom2025HelpfulHH}. Embedding interpretability-guided evaluation within assurance pipelines provides an empirical safeguard against such deception, bridging technical transparency with legal accountability. Our framework extends evaluation into agentic and generative systems, establishing metrics to assess transparency, autonomy, and alignment drift for private auditors and certification bodies. As AI systems become more complex and autonomous, the need for governance-aligned interpretability becomes critical for ensuring these systems remain accountable and controllable. Mechanistic interpretability explains what a model does; governance-oriented evaluation ensures that what it does remains accountable to both private oversight mechanisms and regulatory frameworks.

Building on these mechanistic interpretability advances, we now establish the key requirements that reliable XAI metrics must satisfy to serve as effective governance instruments for both private and public oversight.

\section{Key Requirements for Reliable Metrics}
\begin{figure}
    \centering
    \includegraphics[width=0.69\linewidth]{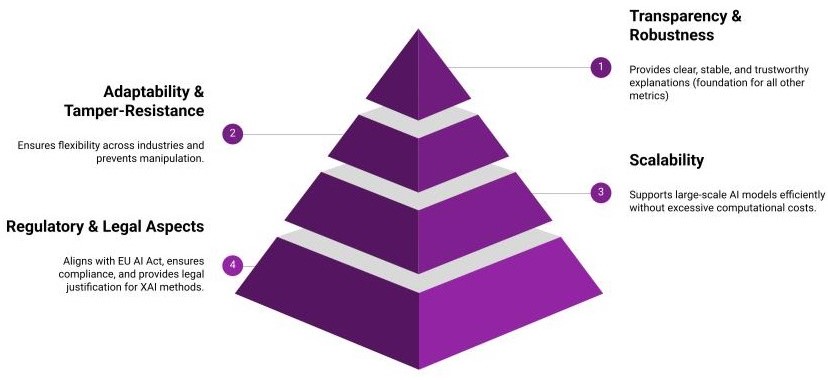}
    \caption{Governance Hierarchy for XAI Evaluation. This pyramid illustrates foundational Transparency \& Robustness, building up through Adaptability \& Tamper-Resistance and Scalability, to Regulatory \& Legal Aspects.}
    \label{fig:governance-hierarchy}
\end{figure}
To overcome these challenges, reliable XAI metrics must be developed. These metrics should establish standard benchmarks for explainability, helping to compare, quantify, and qualify evaluation results. Standardization will clarify regulatory requirements, minimizing bias in user-preferred choices \cite{Nauta_2023,inbff23fook}. As depicted in Figure~\ref{fig:governance-hierarchy}, reliable metrics must meet the following criteria:

\textbf{Transparency \& Robustness : }
XAI evaluation metrics must provide clear, consistent insights into how well an explanation aligns with the model's decision-making process. Transparent metrics ensure stakeholder trust. Metrics should assess the stability of explanations under various conditions, including adversarial inputs and data changes \cite{10.5555/3463952.3463962}. A robust metric ensures that explanations hold up under real-world variations, offering consistency and reliability across different environments. For example, a private auditor evaluating a credit scoring model would require transparent metrics showing not only which features influenced decisions but also how stable those influences remain across different applicant populations and temporal periods.

\textbf{Adaptability \& Tamper-Resistance : }
XAI frameworks should be flexible enough to cater to diverse domains, such as healthcare or finance, where priorities like interpretability or compliance vary \cite{10.5555/3463952.3463962}. Adaptable metrics ensure their applicability across different sectors, addressing the specific challenges of each. Moreover, these metrics must incorporate safeguards such as adversarial testing and regular validation to prevent manipulation, ensuring that explanations are not artificially adjusted to meet predetermined standards. A healthcare certification body, for instance, must ensure that explainability metrics for diagnostic AI adapt to clinical workflows while remaining resistant to manipulation through hyperparameter tuning a balance essential for both usability and integrity.

\textbf{Scalability : }
Scalable explainability metrics are critical for evaluating modern AI systems, particularly large-scale models such as LLMs. Existing explainability metrics are often computationally expensive, limiting their feasibility for large-scale deployments. As AI models grow in complexity, it is imperative to develop scalable methods that can provide meaningful explanations without excessive computational overhead \cite{inprocefr3frfedings}. Certification platforms can batch-evaluate hundreds of models using scalable explainability benchmarks, enabling efficient assessment across diverse model types and sizes, from small specialized models to large foundation models, without prohibitive computational costs.

\textbf{Regulatory \& Legal Aspects : }
While the EU AI Act formalizes transparency and accountability obligations, parallel governance frameworks such as NIST's AI Risk Management Framework (2023) \cite{nist2023ai}, ISO/IEC 42001 (2024) \cite{iso42001}, and Singapore's Model AI Governance Framework \cite{singapore2019} embody equivalent principles of auditability and traceability. XAI metrics must align with regulatory frameworks such as the EU AI Act, which mandates transparency and interpretability for high-risk AI systems \cite{j5010010,10489989}. These metrics must help organizations demonstrate compliance by providing clear, traceable, and auditable explanations. Cross-jurisdictional audit value is enhanced when standardized metrics enable consistent evaluation across different regulatory regimes, facilitating international AI governance cooperation. From a legal perspective, XAI metrics must ensure that AI decisions are explainable and justifiable in court \cite{inproceedingscomplainaceeuract}, particularly in healthcare and finance sectors where AI decisions have significant consequences \cite{Fresz_2024,bibal2020impactlegalrequirementsexplainability}.

Together, these criteria define a governance hierarchy that connects technical transparency to regulatory verifiability.
These requirements inform our governance-by-design approach, which we detail in the following section.

\section{Alternative Views}
Needs differ by domain (e.g., clinical vs. financial), so one-size-fits-all metrics can hinder effectiveness \cite{inprocbhibijbijeedings}. Expert qualitative evaluation complements quantitative metrics to capture context and nuance \cite{r3gr3eginproceedings,colin2023ipredictiunderstand}. A hybrid approach combines core benchmarks (e.g., fidelity, robustness) with domain-specific metrics and human input \cite{ma2024humancentereddesignexplainableartificial}.

Industry stakeholders emphasize practical implementation challenges of private governance mechanisms. Technology companies advocate for flexible, market-driven approaches that allow innovation while maintaining accountability \cite{ball2025frameworkprivategovernancefrontier}. Insurance providers and certification bodies highlight the need for standardized metrics that enable risk assessment and liability determination across diverse AI applications \cite{weil2024insuringemergingrisksai, lior2023innovatingliability}. This diversity of perspectives underscores the importance of adaptable evaluation frameworks that can accommodate sector-specific requirements while maintaining core governance principles.

Governance approaches vary significantly across jurisdictions, reflecting different cultural and legal traditions. While the EU emphasizes prescriptive regulatory frameworks, other regions favor principles-based approaches that rely more heavily on private governance mechanisms \cite{singapore2019}. This variation creates both challenges and opportunities: challenges in achieving cross-border interoperability, but opportunities for regulatory experimentation and learning. Our governance-by-metrics approach provides a common technical foundation that can adapt to diverse regulatory contexts while maintaining consistent evaluation standards.

Our framework acknowledges these diverse perspectives by positioning XAI metrics as flexible governance instruments rather than rigid compliance checklists. The hierarchical model (Figure~\ref{fig:governance-hierarchy}) accommodates domain-specific adaptations while ensuring core requirements transparency, robustness, tamper-resistance remain non-negotiable. Nonetheless, over-standardization may stifle methodological innovation if not balanced with domain flexibility. This balance between standardization and flexibility enables both private governance innovation and regulatory compliance.

Having established these requirements, we now present a comprehensive roadmap for implementing governance-by-metrics in practice.

\section{Governance-by-Design: A Three-Phase Roadmap}

To address the current gaps in XAI evaluation and ensure the development of reliable, transparent, and compliant AI systems, we propose a three-phase governance roadmap that embeds accountability into the evaluation process:

\begin{enumerate}
    \item \textbf{Phase I: Metric Integrity} Develop transparent, reproducible faithfulness metrics and a public integrity registry, covering fidelity, robustness, clarity, and comprehensibility.
    \item \textbf{Phase II: Private Assurance \& Certification} Embed metrics into audits, insurance workflows, and certification processes, enabling private governance actors to assess and verify AI system trustworthiness.
    \item \textbf{Phase III: Regulatory Interoperability} Align private metrics with legal frameworks to ensure traceable compliance across jurisdictions, scalable to advanced models including LLMs, while maintaining compatibility with private governance mechanisms.
\end{enumerate}

Sectors require tailored emphases (e.g., clinical confidence vs. financial compliance and auditability). Explainability methods can be computationally heavy, limiting use on large models. Evaluation must balance efficiency with fidelity to keep interpretability feasible at scale \cite{JeanQuartier2023TheCO}. Collaboration among academia, industry, and regulators is vital to align metrics with real-world needs, especially in high-risk areas \cite{kclBeyondSilos}.

For instance, assurance bodies could use standardized explainability metrics to certify model transparency, enabling interoperability across markets while maintaining competitive flexibility. This phased roadmap is among the first to operationalize governance principles through measurable XAI metrics, enabling continuous accountability for agentic and generative AI systems.

This roadmap's success depends on effective integration with broader policy frameworks and private governance mechanisms. We now examine how XAI metrics can be embedded within existing and emerging governance structures.

\section{Policy Integration: Private and Public Governance Mechanisms}
\begin{figure}[pt]
    \centering
    \includegraphics[width=0.69\linewidth]{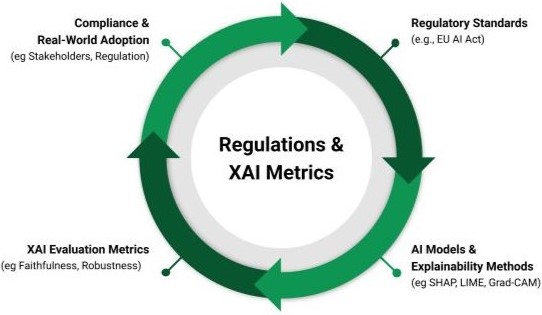}
    \caption{The Iterative Feedback Loop in XAI Regulation and Governance Compliance. Regulatory standards influence XAI evaluation metrics and AI models; subsequent real-world adoption and challenges then refine these standards and metrics to align with evolving AI capabilities.}
    \label{fig:fig2}
\end{figure}

XAI metrics serve as critical instruments for both private and public governance, enabling accountability, trust, and compliance across diverse institutional contexts.

\subsection{Private Governance Mechanisms}

\textbf{Third-Party Auditing and Certification:} Metrics quantify explanation quality for auditors and evaluators \cite{Li2024-bv,McCormack2024-zw}, enabling independent verification by private auditors and certification bodies. Standardized evaluation frameworks allow these actors to assess model trustworthiness across different domains and use cases. For instance, an AI certification body might use standardized explainability metrics to assess whether a healthcare diagnostic model meets transparency requirements before hospital procurement. The certification process would evaluate faithfulness scores, robustness under distribution shift, and consistency across patient demographics providing hospitals with verifiable assurance of model trustworthiness.

\textbf{Insurance and Liability Assessment:} Private governance leverages metrics as market instruments through certification schemes, transparency indices, and risk-adjusted insurance models. These mechanisms enable self-regulation that complements formal oversight, promoting responsible AI through economic and reputational incentives. Consider an AI liability insurance policy: insurers could use explainability metrics to assess risk exposure, offering lower premiums for models demonstrating high faithfulness and robustness scores. This creates market incentives for developers to prioritize explainability, complementing regulatory mandates with economic drivers for responsible AI development.

\textbf{Procurement-Based Governance:} Objective criteria help organizations demonstrate compliance and foster trustworthy deployment \cite{europaRegulation20241689}, enabling procurement-based governance where purchasing decisions incorporate explainability requirements.

\subsection{Regulatory Integration}

\textbf{Regulatory Compliance:} XAI metrics align with legal frameworks (EU AI Act, NIST AI RMF, ISO/IEC 42001) to create auditable templates that satisfy both technical and legal standards.

\textbf{Auditability and Traceability:} Metrics provide clear, consistent evaluation that improves user confidence, especially in high-stakes domains \cite{Sovrano_2021}, while ensuring legal traceability and accountability.

\textbf{Building Trust:} Prioritizing fairness, robustness, and transparency aligns XAI with societal values \cite{Nannini2024-zq,Akhtar2024-gb}, fostering public confidence in AI systems.

\subsection{GPAI Oversight and Cross-Jurisdictional Governance}

As governance frameworks evolve, oversight of General Purpose AI (GPAI) models will depend increasingly on measurable interpretability and alignment metrics. Private governance actors including insurers, AI assurance labs, and model marketplaces play crucial roles in GPAI evaluation, suggesting a need for common explainability metrics that transcend legal boundaries while remaining auditable. For example, a foundation model marketplace might require explainability scores as part of model listings, enabling downstream users to assess transparency before integration. An AI assurance lab could provide independent verification of these scores, creating a trust infrastructure for GPAI deployment that transcends individual regulatory jurisdictions. Integrating interpretability-based alignment into private governance infrastructures provides a safeguard against alignment faking ensuring that models not only appear compliant but demonstrably behave in alignment with ethical and legal standards, providing scalable, jurisdiction-agnostic oversight that complements formal regulation.

\subsection{Broader Societal Implications}

\textbf{Advancing Ethical AI:} Standardized benchmarks enable comparison, best practices, and faster progress. Regulation and explainability co-evolve (Figure~\ref{fig:fig2}), creating a virtuous cycle where technical advances inform governance and governance requirements drive technical innovation.

\section{Conclusion}

Reliable explainability is the foundation of both public and private algorithmic governance. By integrating XAI metrics with global frameworks NIST AI RMF~\cite{nist2023ai}, ISO/IEC 42001~\cite{iso42001}, OECD AI Principles~\cite{oecd2019ai}, and Singapore's Model AI Governance Framework~\cite{singapore2019} this work defines a path toward measurable, auditable accountability. The Governance-by-Metrics paradigm thus extends explainability research into the domain of \emph{AI governance engineering} embedding transparency, accountability, and legal verifiability into the very fabric of model evaluation.

Interpretability-guided alignment demonstrates that reliable explanations serve not only as diagnostic tools but as behavioral governance instruments for GPAI systems, reinforcing the central role of XAI metrics in both private assurance and regulatory compliance. Despite progress in explainability methods, evaluation remains fragmented, subjective, and prone to manipulation. Key challenges include lack of standardization, manipulation risks, limited multi-modal support, and regulatory misalignment. Without rigorous evaluation, explainability risks becoming a mere regulatory formality. Advancing beyond theoretical explainability toward standardized, governance-integrated evaluation frameworks will ensure AI systems remain both accountable and compliant across jurisdictions.

\section{Impact Statement}

Reliable XAI metrics enable cost-efficient auditing, certification, and insurance underwriting for AI systems. By embedding interpretability within private governance infrastructures, these metrics ensure scalable, jurisdiction-agnostic oversight of General Purpose AI. Integrating alignment and auditability within evaluation transforms explainability into an operational mechanism for responsible AI deployment.

\bibliographystyle{unsrt}
\bibliography{references}
\end{document}